# ADVANTAGES IN USING A STOCK SPRING SELECTION TOOL THAT MANAGES THE UNCERTAINTY OF THE DESIGNER REQUIREMENTS

Manuel PAREDES, Marc SARTOR, Cédric MASCLET
LGMT, INSA, 135 avenue de Rangueil 31077 Toulouse Cedex 4
manuel.paredes@ insa-tlse.fr

**Abstract.** This paper analyses the advantages of using a stock spring selection tool that manages the uncertainty of designer requirements. Firstly, the manual search and its main drawbacks are described. Then a computer assisted stock spring selection tool is presented which performs all necessary calculations to extract the most suitable spring from within a database. The algorithm analyses data set with interval values using both multi-criteria analysis and fuzzy logic. Two examples, comparing manual and assisted search, are presented. They show not only that the results are significantly better using the assisted search but it helps designers to detail easily and precisely their specifications and thus increase design process flexibility.

## 1. Introduction

The creation of mechanical objects is often the end result of a long design process. Standard component selection is perhaps the simplest, but nonetheless an important, class of design decision problems as catalogues are becoming increasingly common and voluminous.

Let us analyze the method commonly used by design engineers to select stock springs in order to highlight the difficulties they encounter, the help they can find today and what could be added to improve it.

The usual method for selecting stock springs can be divided in three steps:
- Step 1 : evaluate, from the requirements, certain spring design parameters among those classified in the catalogue of the chosen spring manufacturer.
- Step 2 : find springs that are within parameter limits.
- Step 3 : calculate the operating parameters for each spring short-listed, so as to select one that satisfies the specifications.

Designers are confronted here with the following problem. On the one hand, when the specification is vague, it is difficult to choose the best spring from the large range available. When the specification is precise, on the other hand, choice of an appropriate spring becomes limited.

As paper-based methods are tedious and time-consuming, Yuyi [1] has implemented the first two steps of the search in an expert system but where step 3 has to be calculated. Technical literature provides mathematical methods to calculate the design parameters corresponding to the optimal design (Sandgren [2],

Kannan and Kramer [3], Deb and Goyal [4]). These methods can be used in step 1 but problems have been simplified and the practical existence of the spring is never envisaged.

Text-only systems such as "SPEC" [5] have been developed which assist the designer during step 2. As with many computer-based methods, if the requirements are not specified precisely enough or lie outside the catalogue range, "all or nothing" search results can be obtained [6].

Finally, industrial software available for a designer during the spring definition work such as "Compression spring Software" from IST® [7], can be used in step 3.

The first drawback of the common component selection method is that it may take a long time when the requirements are advanced. The process time remains significant even when dealing with the simplest applications. The second disadvantage is the approximate nature of the procedure. Usually, a spring is selected without being sure of the pertinence of the choice. Generally, the designer cannot affirm that a spring best matching his specifications does not exist.

But the main drawback is probably the decrease in the design process flexibility as illustrated in Fig. 1. Once a stock spring has been selected, the designer is inclined to keep it unchanged for as long as possible to avoid the time-consuming work of making a new search. The choice is usually kept until the end of the design process. So, there is a need for software tools dedicated to reliable and less time-consuming catalogue search.

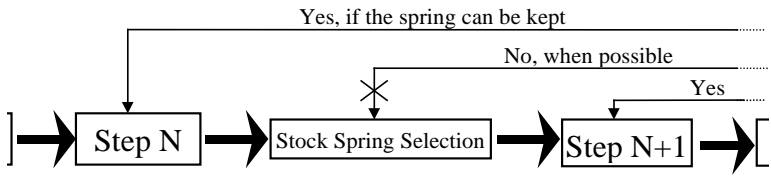

Fig. 1 *Effects on the design process*

This paper presents a tool armed at meeting designer expectations which also takes advantage of all specifications possibilities. The following capabilities, that are not usually presented, are proposed :
- taking uncertain parameters into consideration (data set with interval values),
- automatically performing all the necessary calculations (buckling, fatigue life...) to check that a spring satisfies the specifications,
- excluding "all or nothing" search results,
- introducing an objective function in order to propose the most suitable component.

This tool may be used even in the early design stages. To fit perfectly with the designer's incomplete knowledge, the method determines springs from a specification sheet where data can be uncertain. The associated algorithms select the best spring by calculating the operating parameters for a given objective. Using this kind of tool, the designer can express his specifications in a very formal and practical way. He can obtain search results instantaneously (number of springs available and the one selected). The present study deals only with helical compression springs with closed ends and with closed and ground ends.

## 2. A Stock Spring Selection Tool Working from Toleranced Specifications

First, spring characteristics are detailed in order to illustrate whether the tool can accept over-definite requirements. Then the resolution algorithm is described. Finally, the two different methods available to compare competing springs are presented.

### 2.1. Specifications for the Assisted Search

The parameters which define the spring geometry are: $D_o$, $D$, $D_i$, $d$, $R$, $L_0$, $L_s$, $n$, $z$, $p$. Fig. 2 illustrates these parameters which characterize the intrinsic properties of the spring. A spring works traditionally between two configurations, one corresponding to the least compressed state W1, the other corresponding to the most compressed state W2. Thus the operating parameters which define the use of a spring are: $P_1$, $P_2$, $L_1$, $L_2$ and $sh$ (see
Fig. 3).

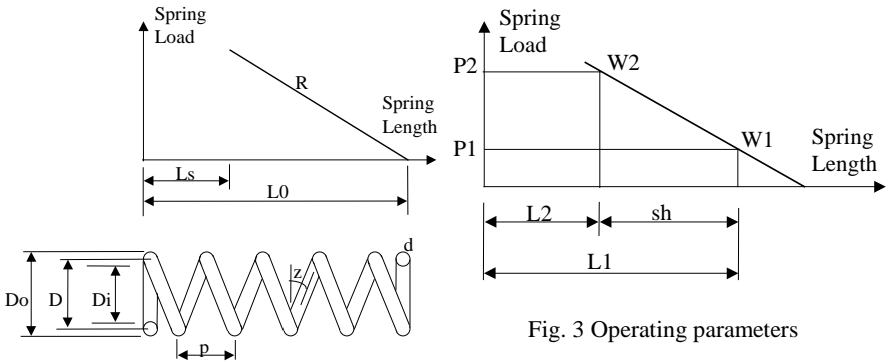

Fig. 3 Operating parameters

Fig. 2 Design parameters

Four independent design parameters have to be known in order to calculate the six others. When the design parameters are known, only two independent operating parameters (to be taken among $P_1$, $P_2$, $L_1$, $L_2$ and $sh$) are necessary to determine the two operating points W1 and W2. To express all the calculations inside the tool which is presented below, springs of the database are defined by $D_o$, $d$, $L_0$ and $R$ and the chosen operating parameters are $L_1$ and $L_2$. Nevertheless, each particular parameter illustrated in Fig. 2 or Fig. 3 can be used by the designer to express his specifications.

The designer can decide on design and operating parameters by giving their bounds (lower and/or upper limits: $L0_S^L$, $L0_S^U$,... $P1_S^L$, $P1_S^U$ ...) in the specification sheet (see Fig. 4). Each fixed parameter simply involves the specification of lower limit equal to upper limit. Moreover, the designer can define a number of other characteristics with interval values:
- Natural Frequency of surge waves
- Spring mass
- Overall space taken up when uncompressed (L=L0)
- Overall space taken up when compressed (L=L2).
- Internal energy during the operating travel

Designers can provide additional data to calculate other characteristics:
- The number of cycles (Ncycles) to calculate the fatigue life factor (to check that it is higher than unity)
- The end fixation factor (ν) to calculate the buckling length and check that it is less than L2. [8]

The designer can also specify the material and the spring ends required.
Finally, to be able to select the most suitable spring, the objective function (maximize fatigue life, minimize mass, minimize L2...) has to be given.

Fig. 4 Specification sheet

Any data not defined in the user specification sheet is set to a default value : 0 for a lower limit and $10^7$ for an upper limit. Then the proposed resolution algorithm is performed.

## 2.2. Proposed Resolution Algorithm

General methods dedicated to component selection problems, as the one proposed by Bradley and Agogino [9], could be applied to the stock spring selection problem. Significant reduction in development costs and processing times can be obtained using a more direct method which is able to take advantage of the spring problem characteristics such as the COSAC system [10] developed at Bath University.

The method chosen here can be considered as the most reliable, since all springs are successively tested. The first spring of the catalogue is evaluated and set as the potential optimum. Then the second spring is evaluated and compared to it. If it is better, it becomes the new potential optimal spring. All the springs of the catalogue are thus evaluated and compared to the last potential optimal spring. When the end of the catalogue is reached, the spring that best matches the requirements is the potential optimal spring. This method provides an acceptable processing time (less than ten seconds) with the catalogue used in this paper which contains about five thousands references.

To evaluate a spring, its four associated design parameters are read from the database and the two operating parameters are automatically calculated in order to optimize the objective value (maximize P2, minimize P2, maximize L2 …) [11]. All previously detailed design and operating parameters are then calculated. To fit with real-life industrial problems, other properties are added such as fatigue life, price, mass, buckling length or solid length. 23 criterions are thus calculated.

When all the spring criteria have been calculated, it remains to know how they fit the specifications. To manage the various needs of designers, two different analysis are proposed.

## 2.3. Comparing Springs Using Multi-criteria Analysis

In the first steps of the design cycle, when most part and shapes have not been chosen, specifications are often imprecise and constraint violations can be admitted. Multi-criteria analysis has been chosen to solve this problem. For each spring, the following equation is used to evaluate the constraint violations.

$$Violation = \frac{\sum_{c=1}^{Nc} K_c \times (Mark_c)}{\sum_{c=1}^{Nc} K_c}$$

The weighting coefficient $K_c$ enables the relative influence of criteria to be adjusted. The mark for criterion c : $Mark_c$ ($L_c$, $U_c$, $V_c$) is calculated as follows :

$L_c$ = Lower bound value of the specifications for criterion c (positive value)
$U_c$ = Upper bound value of the specifications for criterion c
$V_c$ = Criterion value of the spring

$Mark_c = 0$

```
IF V_c > U_c THEN
        IF U_c = 0 THEN
                Mark = V_c
        ELSE
                Mark = (V_c – U_c) / U_c
        END IF
END IF
IF V_c < L_c THEN
        Mark = (L_c – V_c) / L_c
END IF
```

To evaluate a spring, both objective function value (*Objective*) and constraints violation (*Violation*) values have to be taken into account.
The following equation has been selected.

$$Evaluation = Objective \times e^{a \times b \times Violation}$$

where

$a = 1$ if the objective function has to be minimized or $a = -1$ if the objective function has to be maximized.

b is the weighting violation coefficient, in our study, b = 100.

All the springs are then evaluated and the most interesting one according to the *Evaluation* value is selected.

## 2.4. Comparing Springs Using the Fuzzy Logic Analysis

The previous analysis often ends in proposing a spring that is close to certain limit values of the specifications. In the first steps of the design cycle, a spring near the centre of the solution domain can be the best choice, even if its objective function value is less interesting. To solve this kind of problem, fuzzy logic analysis is proposed here.

As the goal is to find springs within the limit values of the specifications, a basic comparison is first carried out on the number of constraints violated (*ncv*). A spring that has the lowest *ncv* value is definitely considered has better than the others. When both springs have the same *ncv* value, the comparison with fuzzy logic is performed. The first step is to evaluate how the tested spring matches the specifications. The comparison between this evaluation and that obtained for the potential optimal spring is made at the second step. In the third step, the comparison between the objective function values of these two springs is done. Then, a final comparison is made to select the new potential optimal spring.

*2.4.1. Step 1 : How a Spring Matches the Specifications*

First, an evaluation is made of each criterion. To evaluate how a criterion matches the specifications, the following method is used.

FuzzyMark_c (L_c, U_c, V_c) = [VB, B, M, G, VG]_c is calculated as follows.

```
IF U_c = 0 THEN
        MarkU_c = - V_c
ELSE
        MarkU_c = (U_c - V_c) / U_c * 100
END IF

MarkL_c = (V_c - L_c) / L_c * 100

WorstMark_c = Min(MarkU_c, MarkL_c)
```

WorstMark$_c$ is used to perform the evaluation of [VB, B, M, G, VG]$_c$

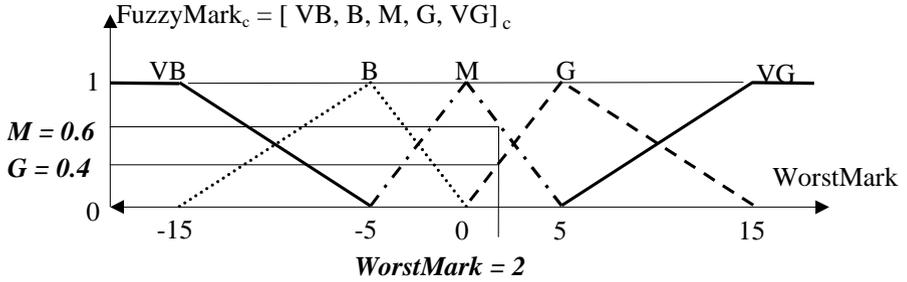

Fig. 5 FuzzyMark$_c$ evaluation

Then, the final values of VB (very bad), B (bad), M (medium), G (good), and VG (very good) are calculated using the formula :

$$[VB, B, M, G, VG]_{Spec} = \frac{\sum_{c=1}^{Nc} K_c \times FuzzyMark_c}{\sum_{c=1}^{Nc} K_c}$$

*2.4.2. Step 2 : Comparison of Specifications*

The comparison of the previous fuzzy values is performed using the Mamdani [12] definition for the AND connector as described in table 1.

Table 1 : Comparison of specifications

|  | Potential optimal spring | | | | |
| --- | --- | --- | --- | --- | --- |
| **Tested spring** | **VB (0)** | **B (0.50)** | **M (0.50)** | **G (0)** | **VG (0)** |
| **VB (0)** | E (0) | S (0) | VS (0) | VS (0) | VS (0) |
| **B (0)** | I (0) | E (0) | S (0) | VS (0) | VS (0) |
| **M (0.70)** | VI (0) | I (0.50) | E (0.50) | S (0) | VS (0) |
| **G (0.30)** | VI (0) | VI (0.30) | I (0.30) | E (0) | S (0) |
| **VG (0)** | VI (0) | VI (0) | VI (0) | I (0) | E (0) |

Then the Or connector (Mamdani) is used to obtain the value of

[VI (very inferior), I (inferior), E (equal), S (superior), VS (very superior)]$_{Spec}$.
Results are shown in table 2.

Table 2 : value of [VI, I, E, S, VS]$_{Spec}$

| VI | I | E | S | VS |
|---|---|---|---|---|
| 0.3 | 0.5 | 0.5 | 0 | 0 |

### 2.4.3. Step 3 : Comparison of Objectives

The comparison of the objective function values is made using the *ObjMark* value:
*Objtop* is the objective function value of the potential optimal spring.
*Objective* is the objective function value of the evaluated spring.
*ObjMark* = 200 * ( *Objtop* - *Objective* ) / ( *Objtop* + *Objective* )

Then the comparison of the two objectives to calculate [VI, I, E, S, VS]$_{Obj}$ is made using the rule defined on fig 6.

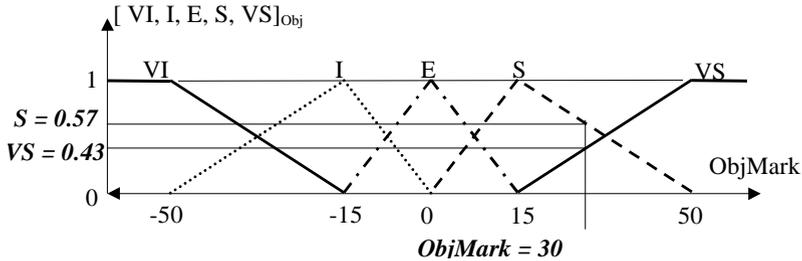

Fig. 6 [VI, I, E, S, VS]$_{Obj}$ value

### 2.4.4. Step 4 : Final Comparison

Springs are finally compared using both [VI, I, E, S, VS]$_{Obj}$ and [VI, I, E, S, VS]$_{Spec}$ with the same rules as in step 2 in order to calculate I, E, S values.

Table 3 : Final comparison

| | [VI, I, E, S, VS]$_{Spec}$ | | | | |
|---|---|---|---|---|---|
| [VI, I, E, S, VS]$_{Obj}$ | VI (0.3) | I (0.50) | E (0.50) | S (0) | VS (0) |
| **VI (0)** | I (0) | I (0) | I (0) | I (0) | E (0) |
| **I (0)** | I (0) | I (0) | I (0) | E (0) | S (0) |
| **E (0)** | I (0) | I (0) | E (0) | S (0) | S (0) |
| **S (0.57)** | I (0.30) | E (0.50) | S (0.50) | S (0) | S (0) |
| **VS (0.43)** | E (0.30) | S (0.43) | VI (0) | S (0) | S (0) |

Table 4 : Value of I, E, S

| I | E | S |
|---|---|---|
| 0.30 | 0.50 | 0.43 |

For the case presented in the previous tables, as S value is superior to I value, the "old" potential optimal spring is superior to the tested spring. Thus, it is kept as the potential optimal one (otherwise the tested spring would have replaced it) and the next spring is tested.

# 3. Examples and Comparison Between Manual and Assisted Search

In order to compare the results between the manual and assisted search, the two examples have been performed using the same catalogue with both methods. Each assisted search has been carried out with the two different analysis. The average time of the first analysis is 3 seconds increasing to 6 seconds for the analysis using fuzzy logic.

## 3.1. A Spring for a Clamping Pin

*3.1.1. Manual Search*

A clamping pin has been custom-designed for an industrial manipulating robot. The time-consuming manual procedure resulted in the selection of the spring Do = 36.0 mm, d = 2.5 mm, L0 = 50 mm, R = 3.54 N/mm, L1 = 47 mm, **L2 = 36 mm**. (steel with closed and ground ends) from the paper catalogue. The result is shown in Fig. 7 and the clamping pin was added to the robot.

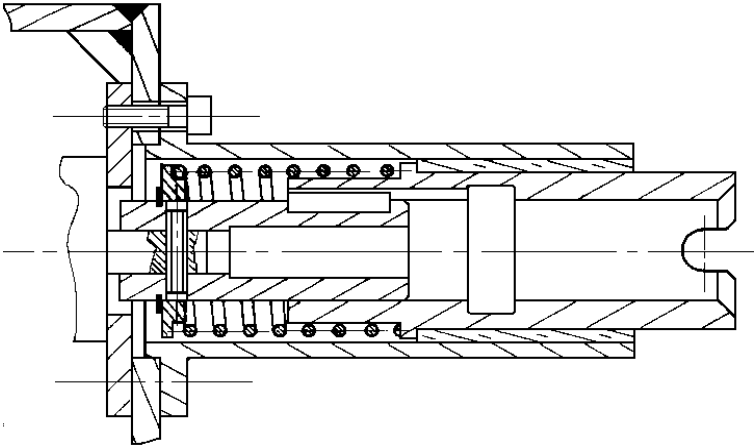

Fig. 7 Old clamping pin design

*3.1.2. Assisted Search*

During a reengineering procedure, it was decided to reselect a spring in order to reduce the main drawback of the clamping pin : its high axial length. Now, the proposed search tool can be used and the specifications can be detailed precisely.

**Specifications :** the maximum outside diameter of the spring is 38 mm, the minimum inside diameter is 27 mm, spring travel must be 11 mm, the maximum value of L1 is 50 mm, the maximum spring rate is 5.5 N/mm, the load P1 must be between 5 and 15N and the load P2 between 50 and 100N. The goal is to obtain the spring with the smallest value of L2.

According to the chosen objective, the algorithm calculates the operating parameters of each spring in order to have the minimum operating length L2 while satisfying the specifications.

**Results :** there are 7 springs that fit the given specifications.
The first analysis proposes a spring that is close to specification requirements (in terms of R) :
Do = 32.0 mm, d = 2.2 mm, L0 = 25 mm, R = 5.78 N/mm, L1 = 22.4 mm, **L2 = 11.4 mm**. (steel with closed and ground ends).
Using fuzzy logic the following spring is selected :
Do = 32.0 mm, d = 2.2 mm, L0 = 32 mm, R = 4.34 N/mm, L1 = 28.54 mm, **L2 = 17.54 mm**. (steel with closed and ground ends).

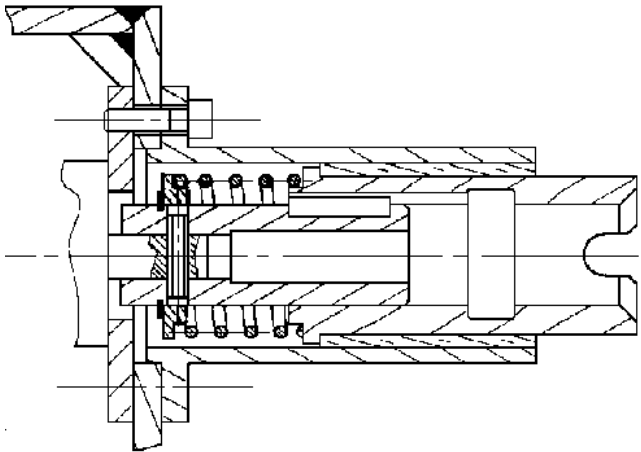

Fig. 8  New clamping pin design

As the properties of the two proposed springs are automatically calculated, the designer can easily choose the one that best matches his specifications. In fact, the spring found with the first method fits the geometrical parameters and has a law *Violation* value (= 0.051), whereas its L2 value is much lower than the one selected by the fuzzy logic method. This spring is chosen for the new design described in Fig. 8. The assisted search proposes springs significantly better than those found with the paper search, leading to useful changes in the clamping pin design.

## 3.2. A Spring for an Axial Displacement Sensor

### 3.2.1. Manual Search
In this example, the manual search in the catalogue led to the following spring :
Do = 12.5 mm, d = 1.25 mm, L0 = 100 mm, R = 0.8 N/mm, L1 = 93.75 mm, L2 = 33.75 mm (**P2 = 53N** , steel with closed and ground ends).

### 3.2.2. Assisted Search
Once again, the assisted search allows to the following requirements to be expressed :
**Specifications :** the maximum outside diameter of the spring is 13 mm, the minimum inside diameter is 5 mm, spring travel must be 60 mm, the minimum P1 value is 3 N and the length L2 must be between 30 and 45 mm. The goal is to obtain the spring with the smallest value of P2.
According to the chosen objective, the algorithm calculates the operating parameters of each spring in order to have the minimum operating load P2 while satisfying the specifications.
**Results :** there are 14 springs that match the given specifications.
Using the first analysis, a spring that fits the specifications (*Violation = 0*) is selected : Do = 11.0 mm, d = 0.9 mm, L0 = 100 mm, R = 0.3 N/mm, L1 = 90 mm, L2 = 30 mm (**P2 = 21N** , steel with closed and ground ends).
Fuzzy logic analysis proposes another spring that fit the specifications :
Do = 11.0 mm, d = 1 mm, L0 = 100 mm, R = 0.374 N/mm, L1 = 92 mm, L2 = 32 mm (**P2 = 25.4N** , stainless steel with closed and ground ends).
In this case, the spring found by fuzzy logic analysis is within the geometrical constraints. In order to obtain a reliable design, this spring is included in the mechanism. Once again, the assisted search results in the choice of a much better spring than the one obtained by the paper based method.

# 4. Conclusion
Paper-based methods for selecting stock springs are tedious, time consuming and decrease design process flexibility. A stock spring selection tool managing uncertain parameters and including all the necessary calculations in order to suggest the most suitable spring is proposed. It has been developed and tested for one year in collaboration with a spring manufacturer. It has shown that this kind of tool changes the designer's approach during the catalogue search. With the assistance of the proposed tool, the designer can specify his needs and quickly choose the spring that best matches his requirements. Finally, this type of tool increases design process flexibility as the component choice is made easier and more efficient.

## Acknowledgements

The financial and technical support of the spring manufacturer « Ressorts VANEL » is gratefully acknowledged.